\RequirePackage{fix-cm}
\documentclass[twocolumn]{svjour3}          % twocolumn
\smartqed  % flush right qed marks, e.g. at end of proof

\usepackage{epsfig}
\usepackage{times}
\usepackage{url}            % simple URL typesetting
\usepackage{booktabs}       % professional-quality tables
\usepackage{amsfonts}       % blackboard math symbols
\usepackage{nicefrac}       % compact symbols for 1/2, etc.
\usepackage{microtype}      % microtypography
\usepackage{xcolor}         % colors
\usepackage{xspace}
\usepackage{graphicx}
\usepackage{wrapfig}
\usepackage{amsmath}
\usepackage{amssymb}
\usepackage{bm}
\usepackage{tabularx}
\usepackage{paralist}
\usepackage{natbib}
\usepackage{bbding}
\usepackage{multirow}
\usepackage[linesnumbered,ruled,vlined]{algorithm2e}
\usepackage{xurl}

\newcommand{\Fref}[1]{Figure~\ref{#1}}

\newcommand{\tref}[1]{Tab.~(\ref{#1})}

\newcommand{\fref}[1]{Fig.~\ref{#1}}

\usepackage{xcolor}
\newcounter{todos}
\AtEndDocument{\ifnum\value{todos}>0 \PackageWarning{TODOS}{There are \arabic{todos} todos left in this paper! Fix them before submitting the paper!} \fi}

\makeatletter
\DeclareRobustCommand\onedot{\futurelet\@let@token\@onedot}
\def\@onedot{\ifx\@let@token.\else.\null\fi\xspace}

\makeatother
\begin{document}
\sloppy
\title{High-Resolution Single-Shot Polarimetric Imaging Made Easy}
% \subtitle{Do you have a subtitle?\\ If so, write it here}


\author{Shuangfan Zhou$^1$ \and Chu Zhou$^2$ \and Heng Guo$^1$ \and  Youwei Lyu$^1$ \and Boxin Shi$^{3,4}$ \and Zhanyu Ma$^1$ \and Imari Sato$^2$}

\authorrunning{Shuangfan Zhou et al.} % if too long for running head

\institute{
\Envelope \, Heng Guo \at
\email{guoheng@bupt.edu.cn} \vspace{0.1cm}\\ 
$1$ \, School of Artificial Intelligence, Beijing University of Posts and Telecommunications, China\\
$2$ \, National Institute of Informatics, Japan\\
$3$ \, State Key Laboratory for Multimedia Information Processing, School of Computer Science, Peking University, China\\
$4$ \, National Key Laboratory of General AI, School of Intelligence Science and Technology, Peking University, China
}

\date{Received: date / Accepted: date}

\maketitle

\begin{abstract}
Polarization-based vision has gained increasing attention for providing richer physical cues beyond RGB images. While achieving single-shot capture is highly desirable for practical applications, existing Division-of-Focal-Plane (DoFP) sensors inherently suffer from reduced spatial resolution and artifacts due to their spatial multiplexing mechanism. To overcome these limitations without sacrificing the snapshot capability, we propose \textbf{EasyPolar}, a multi-view polarimetric imaging framework. Our system is grounded in the physical insight that three independent intensity measurements are sufficient to fully characterize linear polarization. Guided by this, we design a triple-camera setup consisting of three synchronized RGB cameras that capture one unpolarized view and two polarized views with distinct orientations. Building upon this hardware design, we further propose a confidence-guided polarization reconstruction network to address the potential misalignment in multi-view fusion. The network performs multi-modal feature fusion under a confidence-aware physical guidance mechanism, which effectively suppresses warping-induced artifacts and enforces explicit geometric constraints on the solution space. Experimental results demonstrate that our method achieves high-quality results and benefits various downstream tasks.
\end{abstract}

\section{Introduction}
\begin{figure*}[t]
\centering
    \includegraphics[width=1.0\linewidth]{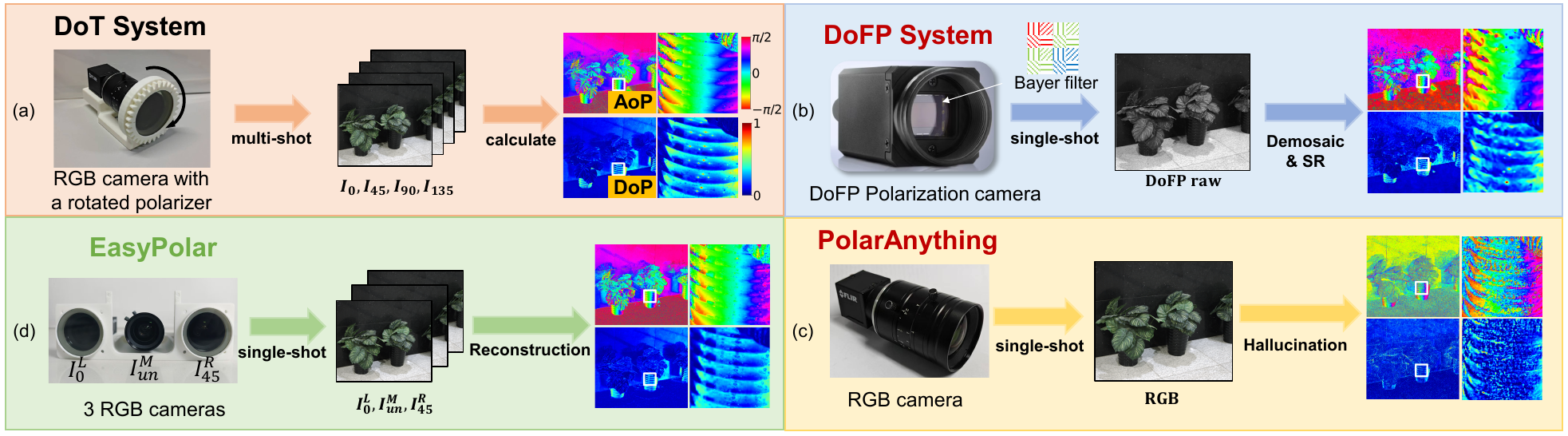}
    \caption{Comparison of representative polarimetric imaging systems. (a) \textbf{DoT system}: a classical Division-of-Time setup that provides high-quality polarization measurements via sequential acquisition, but fails in dynamic scenes due to motion-induced misalignment. (b) \textbf{DoFP camera}: enables snapshot capture through pixel-level multiplexing, at the cost of reduced spatial resolution and demosaicing artifacts. (c) \textbf{PolarAnything}: a generative approach~\citep{zhang2025polaranything} that predicts polarization from a single RGB image, which may lead to physically inconsistent results. (d) \textbf{EasyPolar (ours)}: achieves high-fidelity snapshot imaging by preserving a full-resolution, high-SNR unpolarized reference, avoiding the fundamental resolution and noise trade-offs of DoFP sensors while maintaining physical consistency.}
    \label{fig:Teaser}
    \vspace{-2mm}
\end{figure*}

Polarimetric imaging aims to obtain polarization-related parameterssuch, as the Stokes parameters, Degree of Polarization (DoP), and Angle of Polarization (AoP), by capturing polarized images~\citep{zhou2024quality, zhou2025polarimetric}. These parameters reveal intrinsic light--matter interactions and provide complementary physical cues beyond conventional RGB signals, enabling more robust solutions for downstream vision tasks. Recent progress has demonstrated the benefits of polarization cues in reflection removal~\citep{lyu2019_polarRS, Lei_2020_CVPR}, dehazing in scattering media~\citep{schechner2001instant, zhou2021dehaze}, and glass/transparent object segmentation~\citep{mei2022glass, qiao2023multi}. Moreover, the tight coupling between polarization and surface geometry has been widely exploited in shape-from-polarization to recover fine-grained normals on low-texture or highly reflective surfaces \citep{Lei_2022_CVPR, lyu2024sfpuel, polarfree2025}.

Despite these advantages, acquiring physically accurate and high-resolution polarization data remains a fundamental challenge. Existing polarization capture systems can be broadly categorized into several paradigms. Division-of-Time (DoT) systems, as shown in \fref{fig:Teaser} (a), employ a single RGB camera combined with a rotating polarizer to sequentially record multiple frames. They can produce high-quality polarization measurements, while the scene should be static during the capture. Such multi-shot acquisition inevitably introduces temporal misalignment and motion artifacts in dynamic scenes, leading to inaccurate estimation of polarization-related parameters. Division-of-Focal-Plane (DoFP) sensors based on color-polarization filter arrays (CPFA), as shown in \fref{fig:Teaser} (b), enable single-shot polarization capture by spatially multiplexing polarization states at the pixel level. However, each pixel receives only a fraction of the incident light, leading to a reduced signal-to-noise ratio (SNR) and the absence of a high-SNR unpolarized reference~\citep{zhou2025deblur}. Furthermore, reconstructing full-resolution polarization images requires demosaicing and often super-resolution, which unavoidably introduce interpolation artifacts and distort polarization parameters such as AoP and DoP. The multiplexed sensor design also fundamentally limits the achievable spatial resolution (typically around $2048 \times 2448$ \footnote{Mainstream DoFP sensors, such as the Sony IMX250MZR, utilize a $2 \times 2$ micro-polarizer array that limits the native resolution of each polarization orientation.}).

Beyond these two dominant paradigms, spectroscopic polarization systems split incident light into multiple beams and measure polarization intensities using multiple sensors~\citep{WOLFF199781}. Although capable of accurate capture, such systems are typically complex, bulky, and costly. More recently, learning-based generative approaches have been explored to infer polarization cues directly from RGB images~\citep{zhang2025polaranything}, as illustrated in \fref{fig:Teaser} (c). While attractive for their simplicity, these methods rely on hallucinated predictions without explicit physical measurements, often resulting in physically inconsistent estimates in real-world scenes. \emph{In summary, the complexity, cost, and robustness limitations of existing solutions hinder the widespread adoption of polarimetric imaging in everyday applications.}

To address this problem, we introduce \textbf{EasyPolar}, a new polarimetric imaging framework that addresses the temporal artifacts of DoT systems and the inherent resolution and SNR bottlenecks of DoFP sensors. As shown in \fref{fig:Teaser} (d), we employ three synchronized RGB cameras: one captures a high-SNR unpolarized reference image, while other two are equipped with linear polarizers at distinct orientations to record polarized side-view images. From these observations, EasyPolar reconstructs complete polarization information at the reference view, including all polarization-related parameters derived from the four canonical polarization directions ($0^\circ$, $45^\circ$, $90^\circ$, and $135^\circ$). Compared with conventional DoFP sensors, EasyPolar preserves an unattenuated unpolarized image and avoids the resolution loss caused by pixel-level polarization multiplexing. By leveraging the high quantum efficiency and fabrication maturity of standard RGB cameras, our system enables polarimetric imaging at substantially higher spatial resolution. The modular design of EasyPolar further facilitates extension to multi-sensor platforms, providing a scalable and practical foundation for high-quality polarization capture. 

Building upon the above hardware design, we further propose a confidence-gated polarization reconstruction network to recover physically accurate polarization-related parameters from multi-view observations. The network explicitly integrates geometric priors with a confidence-aware physical guidance mechanism to robustly handle occlusions and warping artifacts inherent to multi-view fusion. Additionally, a pixel-wise PolarEncoder is designed to preserve high-frequency polarization cues, and a direct parameter regression strategy is adopted to ensure physically consistent and high-fidelity reconstruction. We validate the proposed framework using a prototype device built from three synchronized FLIR BFS-U3-123S6C-C RGB cameras\footnote{FLIR BFS-U3-123S6C-C: \url{https://www.teledynevisionsolutions.com/products/blackfly-s-usb3/?model=BFS-U3-123S6C-C}, last accessed on February 10, 2026.} mounted on a fixed multi-view rig with a resolution of $4096 \times 3000$.

To summarize, the main contributions of this paper are:
\begin{itemize}
\item \textbf{A multi-camera polarimetric imaging framework} that captures complete polarization information in a single shot, overcoming the temporal and resolution limitations of existing DoT and DoFP systems.
\item \textbf{A physics-informed reconstruction network} that synergizes explicit geometric priors with confidence-aware physical guidance, enabling robust parameter inference under occlusion and warping.
\item \textbf{A complete system-level validation} through a prototype device and extensive experiments on both reconstruction quality and downstream tasks, demonstrating superiority compared clear advantages over DoFP-based capture.
\end{itemize}

\section{Related Work}
Before introducing our proposed method, we first review the primary polarimetric data acquisition systems and then discuss representative polarization-based applications in downstream low-level vision tasks.
\subsection{Polarimetric Imaging}
\noindent\textbf{DoT systems.} The most classical approach to polarimetric imaging is the DoT system. Such systems obtain polarization information by mechanically rotating a polarizer or using time-multiplexed filter arrays to sequentially capture images at different polarizer angles. This sequential acquisition enables high-fidelity, full-resolution polarization measurements, as each frame is captured without spatial multiplexing. However, it inherently requires multiple exposures, making the process time-consuming and highly susceptible to temporal misalignment in dynamic scenes. Consequently, DoT systems are unsuitable for scenarios that demand high responsiveness, such as imaging from moving platforms or capturing fast-changing environments.

\noindent\textbf{DoFP sensors.} In contrast to DoT systems, DoFP sensors have become the predominant architecture for real-time polarimetric imaging. By integrating a micro-polarizer array (commonly a CPFA) onto the sensor plane, they enable the acquisition of multiple polarized images in a single shot. However, the spatial multiplexing inherent in the sensor design introduces several challenges. First, the raw CPFA output requires a complex demosaicing procedure to reconstruct full-resolution polarized images, which inevitably introduces interpolation artifacts that degrade parameters such as DoP and AoP. Second, DoFP sensors generally offer lower spatial resolution than conventional RGB cameras. Third, they cannot provide a high-SNR unpolarized reference image. Consequently, extensive research has focused on improving DoFP output quality through advanced demosaicing~\citep{qiu2019pid, liu2020pid, lu2024pid, liu2023pid, zheng2024pid} and super-resolution (SR)~\citep{hu2023polarized, yu2023color} techniques. To further enhance overall performance, recent state-of-the-art methods, such as PIDSR~\citep{zhou2025pidsr}, propose unified frameworks that jointly perform demosaicing and SR. Nevertheless, significant limitations remain, leaving ample room for further quality improvement.

\noindent\textbf{Amplitude-splitting systems.} Another category of snapshot polarimetric imaging relies on amplitude-splitting architectures. These systems typically utilize complex optical assemblies, such as polarizing beam splitters (PBS) or prisms, to direct light into multiple optical paths captured by separate sensors~\citep{WOLFF199781}. While this design achieves simultaneous acquisition with full spatial resolution and high SNR, it suffers from significant hardware complexity. Constructing such setups requires specialized components, including quarter-wave plates, relay lenses, and prisms, and demands precise pixel-level optical alignment to register images across different sensors. Such stringent requirements often result in bulky, expensive, and fragile systems that are difficult to deploy in general-purpose.

\noindent\textbf{Generative approaches.} To overcome the limitations of DoFP sensors, generative models have recently emerged as a promising alternative for synthesizing polarization information directly from RGB images~\citep{zhang2025polaranything, lin2025rgb}. By leveraging high-SNR RGB inputs without an on-chip polarizer array, these methods avoid the need for complex demosaicing and super-resolution processes. However, generative approaches are inherently challenged by the nature of synthesis: the predicted polarization cues may be physically inconsistent or lack fine-grained details, potentially leading to inaccuracies when compared with real measured data.

\subsection{Polarization-based Vision Applications}
\noindent\textbf{Shape from polarization.} Shape-from-Polarization (SfP) is a critical technique that reconstructs 3D geometry by leveraging the intrinsic relationship between surface normals and polarization-related parameters (typically the AoP and DoP). Specifically, methods like PANDORA~\citep{dave2022pandora} integrate polarization information into Neural Radiance Fields (NeRFs) to achieve high-fidelity shape reconstruction. Other approaches, such as MVAS \citep{cao2023multi}, propose geometry estimation using only multi-view AoLP inputs, eliminating the dependence on auxiliary RGB imagery. Furthermore, SFPUEL~\citep{lyu2024sfpuel} combines photometric stereo with polarization cues to robustly estimate shapes from a single viewpoint. Collectively, these polarization-aware methods yield demonstrably more accurate and reliable shape reconstructions, establishing SfP as a powerful paradigm in 3D surface analysis. Beyond direct geometry estimation, polarization also plays a crucial role in depth estimation. Several techniques utilize polarization to enhance stereo depth estimation~\citep{Ikemura_2024_CVPR,smith2016linear,zhu2019depth,Fukao_2021_CVPR,tian2023dps}, consistently improving accuracy, particularly in challenging scenarios.

\noindent\textbf{Reflection removal and transparent object segmentation.} Both reflection removal and transparent object segmentation exploit polarization's inherent ability to discriminate between reflected and transmitted light, proving highly effective for the robust decomposition of reflection and transmission layers on glossy or transparent substrates. For reflection removal, Polar-RS~\citep{lyu2019_polarRS} employs a polarized-unpolarized image pair to effectively isolate and decompose reflection layers. Similarly, PolarFree~\citep{polarfree2025} introduces a novel diffusion-based model that achieves state-of-the-art performance across diverse operational environments. In the domain of transparent object segmentation, polarization's sensitivity to partial reflections is critical. Pioneering techniques by~\cite{kalra2020deep} and~\cite{mei2022glass} capitalize on polarization to successfully segment glass from highly specular or cluttered backgrounds in complex robotic and outdoor scenarios. More recently,~\cite{qiao2023multi} extended these principles to video processing, facilitating the real-time segmentation of transparent objects in dynamic scenes.

\noindent\textbf{Image quality enhancement.} Polarization serves as a vital cue for enhancing RGB image fidelity across various challenging domains. The modality's inherent capacity to decouple highly polarized scattered light from the direct, often weakly polarized, target radiance is exploited by methods addressing dehazing~\citep{zhou2021dehaze, schechner2001instant}, descattering~\citep{Treibitz2009descatter, li2025descatter}, and underwater imaging restoration \citep{hu2022underwater, shen2024underwater, su2024underwater}. This principle yields substantial improvements in visibility and contrast across diverse turbid media, including foggy, misty, or submerged environments. Furthermore, leveraging the distinct polarization signatures of scene objects facilitates other challenging image quality restoration problems, including high dynamic range reconstruction~\citep{Ting2021HDR, zhou2023HDR} and robust shadow removal~\citep{zhou2025polarization}.

\section{Preliminary \& Problem Formulation}
The proposed EasyPolar framework aims to recover high-fidelity polarization parameters in a physically consistent manner without sacrificing the efficiency of single-shot acquisition. Unlike conventional approaches that trade spatial resolution or robustness for instantaneous capture, EasyPolar adopts a system-level design that jointly optimizes the sensing configuration and reconstruction strategy. Specifically, we combine a carefully designed multi-view polarization capture scheme with a physics-informed reconstruction network to preserve high-SNR radiance information while enabling complete polarimetric recovery. This section details the physical image formation model, the design principles underlying the proposed sensing configuration, and a formal problem formulation that casts polarization reconstruction as a constrained inference task under cross-view inconsistencies.

\subsection{Physical Image Formation Model}
Polarized images can be captured by placing a linear polarizer at an angle $\alpha$ in front of the camera. By rotating the polarizer to different angles, images under multiple polarization states can be obtained. The intensity measured by the sensor at a given polarizer angle $\alpha$ is modeled as
\begin{equation} 
I_{\alpha} = \frac{I_{\max} + I_{\min}}{2} + \frac{I_{\max} - I_{\min}}{2} \cos 2(\alpha - \theta), 
\end{equation} 
where $I_{\max}$ and $I_{\min}$ denote the maximum and minimum observed intensities, and $\theta$ represents the AoP of the incident light.

From images captured at different polarizer angles, the Stokes vector $\mathbf{S} = [S_0, S_1, S_2]^\top$ can be computed to characterize the polarization state. Since our system uses only linear polarizers, we do not consider the circular polarization component $S_3$. The Stokes parameters are calculated as
\begin{equation}
\left\{
\begin{aligned}
S_0 &= \frac{I_{0} + I_{45} + I_{90} + I_{135}}{2}, \\
S_1 &= I_{0} - I_{90}, \\
S_2 &= I_{45} - I_{135},
\end{aligned}
\right.
\label{eq:stokes}
\end{equation}
where $I_{\alpha}$ represents the intensity captured when the transmission axis of the polarizer is oriented at an angle $\alpha \in \{0^\circ, 45^\circ, 90^\circ, 135^\circ\}$.

Based on the computed Stokes parameters, two key physical quantities can be derived: the AoP $\theta$ and DoP $\rho$. They are given by
\begin{equation} 
\theta = \frac{1}{2} \arctan\left(\frac{S_2}{S_1}\right), \quad \rho = \frac{\sqrt{S_1^2 + S_2^2}}{S_0}.
\label{eq:aop}
\end{equation} 
Note that $S_0$ corresponds to the unpolarized intensity. In the rest of this paper, we denote $S_0$ as $I_{un}$ for clarity. In addition, we introduce the following physical property, which will be used in our method:
\begin{equation}
I_0 + I_{90} = I_{45} + I_{135} = I_{un}.
\label{eq:polar}
\end{equation}

On the other hand, polarimetric signals are intrinsically coupled with surface geometry, specifically the surface normal parameterized by the azimuth angle $\phi$ and the zenith angle $\vartheta$. The observed AoP $\theta$ is related to the azimuth angle $\phi$ with an ambiguity that depends on the dominant reflection mechanism:
\begin{equation}
\theta =
\begin{cases} 
\phi, & \text{if diffuse reflection dominates} \\
\phi - \frac{\pi}{2}, & \text{if specular reflection dominates}
\end{cases}.
\label{eq:azimuth_geo}
\end{equation}

Regarding the zenith angle $\vartheta$, it determines the DoP $\rho$ given the refractive index $n$. When diffuse reflection dominates, the relationship follows the Fresnel equations~\citep{Atkinson2006}:
\begin{equation}
\rho_d = \frac{(n - \frac{1}{n})^2 \sin^2\vartheta}{2 + 2n^2 - (n + \frac{1}{n})^2 \sin^2\vartheta + 4 \cos\vartheta \sqrt{n^2 - \sin^2\vartheta}}.
\label{eq:dolp_diffuse}
\end{equation}
Conversely, when specular reflection dominates, the degree of polarization is derived as:
\begin{equation}
\rho_s = \frac{2 \sin^2\vartheta \cos\vartheta \sqrt{n^2 - \sin^2\vartheta}}{n^2 - \sin^2\vartheta - n^2 \sin^2\vartheta + 2 \sin^4\vartheta}.
\label{eq:dolp_specular}
\end{equation}

\subsection{Sensing Configuration}
Recovering the linear Stokes vector requires at least three independent intensity measurements to resolve the unknowns $S_0, S_1$, and $S_2$. Rather than distributing these measurements uniformly across multiple views, we adopt a hybrid multi-view sensing configuration that explicitly prioritizes the quality of the reference radiance. As illustrated in \fref{fig:Teaser} (d), one viewpoint, denoted as the primary viewpoint ($M$), serves as the reconstruction target and captures the unpolarized intensity $I_{un}$. This observation preserves the complete scene radiance without polarization-induced attenuation, providing a high-SNR and high-frequency spatial reference. To introduce the required polarimetric diversity, two auxiliary viewpoints, left ($L$) and right ($R$), capture linearly polarized intensities at orientations of $0^\circ$ and $45^\circ$, yielding $I_0^L$ and $I_{45}^R$, respectively.

After spatially aligning the auxiliary observations to the primary viewpoint $M$, the linear Stokes components can be expressed as:
\begin{equation}
\label{eq:stokes_inversion}
S_0 = I_{un}^M, \quad S_1 = 2 I_0^M - I_{un}^M, \quad S_2 = 2 I_{45}^M - I_{un}^M.
\end{equation}
The target parameters, namely the AoP $\theta$ and the DoP $\rho$, are then uniquely determined by substituting Eq.~\eqref{eq:stokes_inversion} into Eq.~\eqref{eq:aop}. This configuration ensures polarimetric completeness while anchoring the reconstruction to a high-quality unpolarized reference, thereby mitigating the resolution loss and noise amplification commonly encountered in pixel-level polarization multiplexing

\subsection{Problem Formulation}
Building upon the proposed sensing configuration, our objective is to recover polarization parameters aligned with the primary viewpoint $M$ from hybrid multi-view observations. Formally, we seek to learn a mapping $f_{\Theta}$:
\begin{equation}
\theta^M, \rho^M = f_{\Theta}(I_0^L, I_{un}^M, I_{45}^R),
\label{eq:objective}
\end{equation}
where $\Theta$ denotes the learnable parameters. Although Eq.~\eqref{eq:stokes_inversion} provides a theoretical reconstruction pathway under ideal conditions, the physical observations exhibit inherent coordinate displacements and viewpoint-dependent variations. Our framework resolves these cross-view inconsistencies in an end-to-end manner by jointly reasoning over spatial correspondence, confidence, and physical constraints. By regressing the physical parameters $\theta^M$ and $\rho^M$ directly, the above formulation yields a consistent and physically meaningful polarimetric representation in a single shot, while fully leveraging the complementary information provided by the auxiliary views.

\begin{figure*}[t]
\centering
    \includegraphics[width=1.0\linewidth]{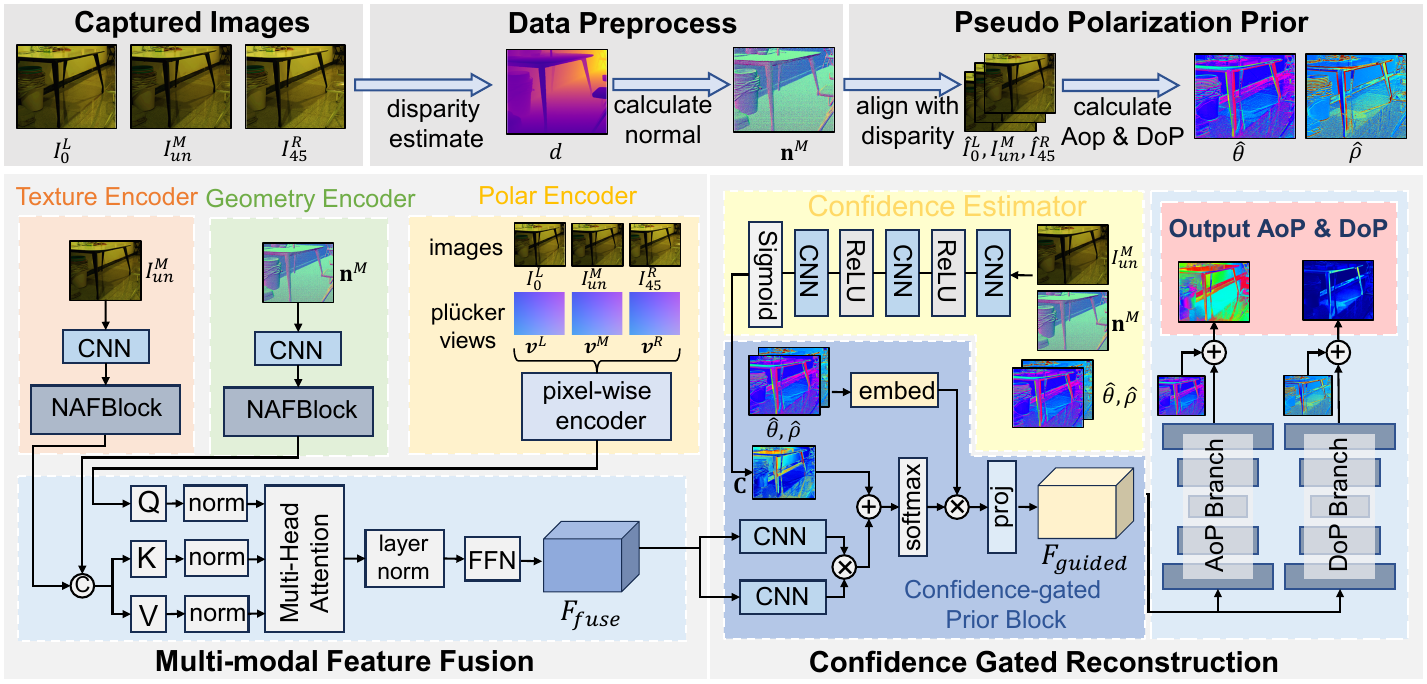}
    \caption{Overview of the proposed physics-informed reconstruction network. Given three input images with distinct viewpoints and polarization states, we first perform viewpoint alignment and surface normal estimation using a pre-trained disparity network. The framework then employs a multi-modal feature fusion module to extract and aggregate features from polarization, texture, and geometry. Subsequently, a confidence-gated reconstruction module uses pseudo-polarization priors to refine the fused features, while adaptively controlling the influence of these priors via an estimated confidence map $\mathbf{C}$. By combining these components, the network leverages physical constraints and confidence guidance to produce high-fidelity AoP and DoP estimates.}
    \label{fig:Network}
    \vspace{-2mm}
\end{figure*}

\section{Physics-Informed Reconstruction Network}
\label{sec:network}

To solve the parameter regression task defined in Eq.~\eqref{eq:objective}, we develop a physics-informed reconstruction network. The network is designed to adaptively aggregate polarimetric and structural cues from the hybrid observations, translating multi-view intensities into a unified, physically consistent polarization representation aligned with the primary viewpoint.

\subsection{Overall Architecture}
The physics-informed reconstruction network is designed to recover the polarization parameters $\theta^M$ and $\rho^M$ from the hybrid observations $\mathcal{I}$. The primary challenge lies in the inherent spatial displacement and viewpoint-dependent intensity variations among $\{I_0^L, I_{un}^M, I_{45}^R\}$. Simple geometric alignment is often insufficient for high-fidelity reconstruction, as warping can introduce occlusion artifacts and fails to account for the physical shifts in polarization states across viewing angles.

To address these challenges, we propose a learning-based architecture that integrates geometric alignment with polarization synthesis. As illustrated in Fig.~\ref{fig:Network}, the network processes the observations through a sequence of modules:

\subsection{Initialization via Pseudo Priors}
To facilitate learning of the mapping $f_\Theta$, we first transform the raw observations into a shared coordinate space at the reference viewpoint. Since $I_0^L$ and $I_{45}^R$ are captured from different spatial positions, we must establish their pixel-wise correspondence with $I_{un}^M$. Specifically, we employ IGEV-Stereo~\citep{xu2023iterative} to estimate dense disparity maps between the side views and the reference view. We denote $d_L$ as the disparity from the left view to the middle view, and $d_R$ as the disparity from the right view to the middle view. Based on these disparities, the side-view observations are aligned to the reference viewpoint through a warping function $\mathcal{W}(\cdot)$:
\begin{equation}
\hat{I}_0^M = \mathcal{W}(I_0^L, d_L), \quad \hat{I}_{45}^M = \mathcal{W}(I_{45}^R, d_R).
\label{eq:warping}
\end{equation}

Based on these, we augment the network input with complementary geometric and polarimetric cues:

\noindent\textbf{Polarimetric pseudo priors.} We analytically derive an initial polarization distribution by substituting the aligned intensities into the Stokes inversion in Eq.~\eqref{eq:stokes_inversion}. We denote $\tilde{\theta}$ and $\tilde{\rho}$ as pseudo priors because they are computed from the warped intensities $\hat{I}_0^M$ and $\hat{I}_{45}^M$. While these estimates provide essential physical cues, they are inherently affected by viewpoint-dependent polarimetric variations and potential inaccuracies introduced during the warping process.

\noindent\textbf{Geometric priors.} We recover scene geometry from stereo correspondence to provide structural context. We compute the metric depth $z^M$ at the reference viewpoint from the estimated disparity $d$:
\begin{equation}
z^M = \frac{f \times B}{d},
\label{eq:depth}
\end{equation}
where $f$ and $B$ denote the focal length and baseline, respectively. Since surface normals are more closely coupled with polarization states than raw depth (see Eqs.~\eqref{eq:azimuth_geo} and~\eqref{eq:dolp_specular}), we further derive the surface normal map $\mathbf{n}^M$ from the spatial gradients of the depth map:
\begin{equation}
\mathbf{n}^M = \frac{(-\frac{\partial z}{\partial x}, -\frac{\partial z}{\partial y}, 1)^\top}{\sqrt{(\frac{\partial z}{\partial x})^2 + (\frac{\partial z}{\partial y})^2 + 1}}.
\label{eq:normal}
\end{equation}
By providing $\mathbf{n}^M$ as additional input, the network can more effectively leverage the intrinsic physical constraints between surface orientation and polarimetric response to refine the initial estimates.

\subsection{Multi-modal Feature Fusion Module}
The first stage of the EasyPolar network is designed to aggregate multi-modal cues and resolve the inherent ambiguities in sparse, single-shot measurements. Given the corrupted nature of the pseudo priors $\tilde{\theta}, \tilde{\rho}$, this stage focuses on establishing a latent representation that harmonizes the high-SNR structural context of the reference view with the angular-dependent polarimetric signals.

\noindent\textbf{Texture and geometry encoding.} 
The unpolarized image $I_{un}^M$ captures intensity textures, while the normal map $\mathbf{n}^M$ represents geometric shape priors. We employ two encoders, $E_{tex}$ and $E_{geo}$, based on the NAFBlock architecture \citep{chen2022simple}, to extract hierarchical features from these inputs:
\begin{equation}
\mathbf{F}_{tex} = E_{tex}(I_{un}^M), \quad \mathbf{F}_{geo} = E_{geo}(\mathbf{n}^M).
\end{equation}

\noindent\textbf{Polarimetric encoding.}
To extract polarimetric features $\mathbf{F}_{pol}$ from the multi-view observations $\{I_0^L, I_{un}^M, I_{45}^R\}$, we adopt a pixel-wise encoding strategy to preserve the integrity of signal differences. The input to this encoder consists of the concatenated intensity images and their corresponding viewing rays $\mathbf{v}^L$, $\mathbf{v}^M$, and $\mathbf{v}^R$ for the left, middle, and right cameras, respectively. Since polarimetric cues are primarily embedded in intensity variations across polarization angles, this encoder uses $1 \times 1$ kernels to maintain high-fidelity pixel-level relationships and prevent standard spatial convolutions from smoothing out these subtle variations. Furthermore, because polarization states are inherently viewpoint-dependent, we incorporate viewing rays to provide geometric context for the encoding process. Each ray $\mathbf{v}$ is parameterized using Plücker coordinates $\mathcal{P} = (\mathbf{d}, \mathbf{m})$, where $\mathbf{d} \in \mathbb{S}^2$ denotes the ray direction derived from camera intrinsics and $\mathbf{m} = \mathbf{o} \times \mathbf{d}$ represents the ray moment determined by the camera center $\mathbf{o}$ in world coordinates. A Feature-wise Linear Modulation (FiLM) module takes these concatenated 18D coordinates as input to generate spatial scaling factors $\gamma$ and shifting offsets $\beta$ to modulate intermediate features. This mechanism enables the network to adaptively compensate for viewpoint-dependent shifts and map the multi-view observations into a view-invariant polarimetric feature space $\mathbf{F}_{pol}$.

\noindent\textbf{Multi-modal fusion.} 
To effectively integrate these diverse features, we utilize a window-based cross-attention mechanism. Since the primary goal is to refine the polarimetric estimate, we treat $\mathbf{F}_{pol}$ as the main Query, while the texture $\mathbf{F}_{tex}$ and geometry $\mathbf{F}_{geo}$ serve as the structural context. We first concatenate the structural features to form a comprehensive context $\mathbf{F}_{ctx} = [\mathbf{F}_{tex}, \mathbf{F}_{geo}]$. Within non-overlapping local windows, the Query ($\mathbf{Q}$), Key ($\mathbf{K}$), and Value ($\mathbf{V}$) projections are computed as:
\begin{equation}
\mathbf{Q} = \mathbf{F}_{pol}\mathbf{W}_Q, \quad \mathbf{K} = \mathbf{F}_{ctx}\mathbf{W}_K, \quad \mathbf{V} = \mathbf{F}_{ctx}\mathbf{W}_V,
\end{equation}
where $\mathbf{W}_Q, \mathbf{W}_K, \mathbf{W}_V$ are learnable linear projection matrices. The fused feature map $\mathbf{F}_{fuse}$ is then obtained by aggregating the context information based on the attention map, followed by a residual connection to preserve the original polarimetric cues:
\begin{equation}
\mathbf{F}_{fuse} = \text{Softmax}\left(\frac{\mathbf{Q}\mathbf{K}^T}{\sqrt{d_k}}\right)\mathbf{V} + \mathbf{F}_{pol},
\end{equation}
where $d_k$ is the scaling factor. By querying the structural context with $\mathbf{F}_{pol}$, the network adaptively enhances the polarimetric representation in regions with rich texture or geometric details.

This formulation enables the network to selectively enhance the polarimetric representation based on local geometric complexity and texture richness. For instance, in regions with complex shading where $\tilde{\theta}$ and $\tilde{\rho}$ are unreliable, the attention mechanism assigns higher weights to the structural context $\mathbf{F}_{ctx}$ to enforce physical consistency. The output of this stage is a comprehensive fused feature map $\mathbf{F}_{fuse}$, which serves as the input for the subsequent synthesis stage.

\subsection{Confidence Gated Reconstruction Module}
The primary goal of this module is to transform the fused multi-modal features $\mathbf{F}_{fuse}$ into high-fidelity polarimetric representations. While the initial pseudo estimates $\tilde{\theta}$ and $\tilde{\rho}$ provide valuable physical cues, they can suffer from artifacts due to warping errors and occlusions. We therefore introduce a confidence-aware mechanism to assess the reliability of these priors and guide the feature learning process.

\noindent\textbf{Confidence estimation.} 
To explicitly identify unreliable regions, we employ a Confidence Network $\mathcal{C}$ to predict a pixel-wise reliability map $\mathbf{C} \in [0, 1]$. This network takes the unpolarized intensity $I_{un}^M$, surface normals $\mathbf{n}^M$, and the pseudo parameters $\tilde{\theta}$ and $\tilde{\rho}$ as input:
\begin{equation}
    \mathbf{C} = \mathcal{C}( I_{un}^M, \mathbf{n}^M, \tilde{\theta}, \tilde{\rho} ).
\end{equation}
A high value in $\mathbf{C}$ indicates that the local pseudo priors are trustworthy, while a low value signals potential errors at occlusion boundaries or misaligned regions.

\noindent\textbf{Confidence gated fusion.} 
We propose a Confidence-Gated Fusion Block to inject physical guidance into the feature space while preventing error propagation. The core idea is to use the pseudo priors to guide $\mathbf{F}_{fuse}$ in learning polarization-specific patterns, while using $\mathbf{C}$ to gate the flow of unreliable information. Specifically, we construct a bias matrix $\mathbf{M}$ derived from the confidence map:
\begin{equation}
    \mathbf{M} = \log(\mathbf{C} + \epsilon),
\end{equation}
where $\epsilon$ is a small constant for numerical stability. This bias $\mathbf{M}$ is added to the attention logits before the Softmax operation. In low-confidence regions where $\mathbf{C}$ approaches zero, $\mathbf{M}$ tends toward $-\infty$, effectively zeroing out the attention weights for unreliable areas. This forces the network to aggregate information solely from trustworthy spatial neighbors, resulting in a guided feature map $\mathbf{F}_{guided}$ that is robust to input artifacts.

\noindent\textbf{Dual-branch reconstruction.} 
The final reconstruction is performed by two independent parallel branches. To preserve the rich multi-modal context from the initial stage, each branch takes both the guided features $\mathbf{F}_{guided}$ and the initial fused features $\mathbf{F}_{fuse}$ as input. The DoP branch directly estimates the degree of linear polarization $\rho^M$. Simultaneously, the AoP branch predicts the continuous components $\sin 2\theta^M$ and $\cos 2\theta^M$ to enforce geometric continuity and avoid phase ambiguity. This targeted regression helps ensure that the output remains physically consistent across the entire scene.

\begin{figure*}[t]
\centering
    \includegraphics[width=1.0\linewidth]{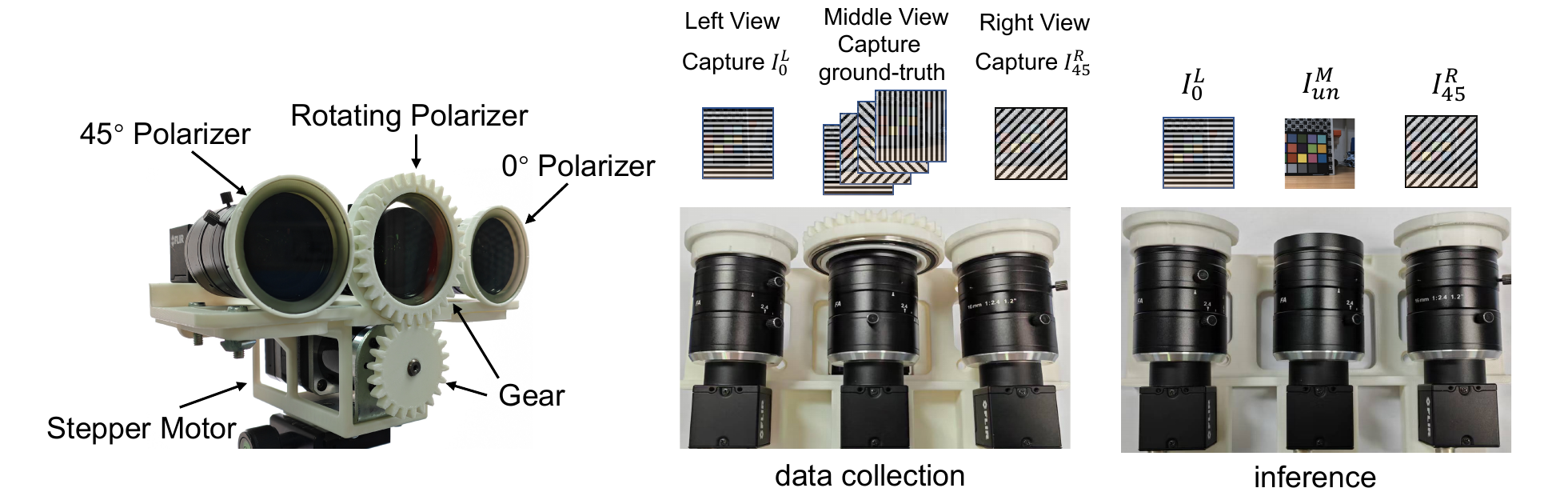}
    \caption{Our real-world data acquisition setup. The prototype comprises three synchronized RGB cameras mounted with a narrow horizontal baseline. The left and right cameras are equipped with fixed linear polarizers oriented at $0^\circ$ and $45^\circ$, respectively. The middle camera temporarily fitted with a motorized rotatable polarizer \emph{only during data collection} to enable time-multiplexed measurements for high-precision ground-truth computation. During inference, this polarizer is removed, and the central view serves as an unpolarized reference for single-shot capture.}
    \label{fig:setup}
\end{figure*}

\subsection{Loss Function Design}
The training is supervised by a composite loss function designed to handle the data imbalance in polarimetric signals and ensure physical consistency. 

\noindent\textit{1) Polarization parameter loss ($\mathcal{L}_P$).}
This term directly supervises the network's output AoP and DoP. For AoP, in addition to the pixel-wise $\ell_1$ loss on the sine and cosine components, we introduce a Gradient Loss ($\mathcal{L}_{grad}$) to preserve geometric structures and edge details. The AoP-related loss is defined as:
\begin{equation}
\begin{split}
\mathcal{L}_{\theta} &= \| \sin(2\hat{\theta}) - \sin(2\theta^{gt}) \|_1 + \| \cos(2\hat{\theta}) - \cos(2\theta^{gt}) \|_1 \\
&+ \lambda_{g} \Big( \| \nabla \sin(2\hat{\theta}) - \nabla \sin(2\theta^{gt}) \|_1 \\
&\quad\quad + \| \nabla \cos(2\hat{\theta}) - \nabla \cos(2\theta^{gt}) \|_1 \Big),
\end{split}
\end{equation}
where $\nabla$ denotes the spatial gradient operator. For DoP, considering that significant polarimetric signals are often sparse in natural scenes, we define a general weighted loss as:
\begin{equation}
\mathcal{L}_{w}(y, y^{gt}, \mathbf{W}) = \frac{1}{N}\sum_{i} \mathbf{W}_i \cdot | y_i - y^{gt}_i |_1.
\label{eq:weighted_l1}
\end{equation}
By assigning the weight map as $\mathbf{W} = 1 + \alpha \cdot y^{gt}$, the DoP loss is formulated as:
\begin{equation}
\mathcal{L}_{\rho} = \mathcal{L}_{w}(\hat{\rho}, \rho^{gt}, \mathbf{W}_{\rho}).
\label{eq:dop_loss}
\end{equation}
 The combined polarization loss is then expressed as:
\begin{equation}
\mathcal{L}_P = \mathcal{L}_{\theta} + \mathcal{L}_{\rho}.
\end{equation}

\noindent\textit{2) Confidence loss ($\mathcal{L}_{conf}$).}
We supervise the predicted confidence map $\hat{\mathbf{C}}$ against the ground truth confidence $\mathbf{C}^{gt}$. $\mathbf{C}^{gt}$ is derived from the physical consistency error between the ground-truth geometric warping of views.
Similar to DoP, we apply a weighted loss to focus on error-prone regions (low confidence). The weight is defined as $\mathbf{W}_{c} = 1 + \beta \cdot (1 - \mathbf{C}^{gt})$.
\begin{equation}
\mathcal{L}_{conf} = \mathcal{L}_{w}(\hat{\mathbf{C}}, \mathbf{C}^{gt}, \mathbf{W}_{c}).
\end{equation}

\noindent\textit{3) Intensity reconstruction loss ($\mathcal{L}_I$).}
To ensure the reconstructed parameters are physically valid, we synthesize the polarized images at $0^\circ$ and $45^\circ$ from the predicted parameters and the unpolarized intensity $I_\text{un}^M$. We then calculate the $\ell_1$ loss against the ground truth images:
\begin{equation}
\mathcal{L}_I = \| \hat{I}_0^M - I_0^M \|_1 + \| \hat{I}_{45}^M - I_{45}^M \|_1.
\end{equation}

\noindent\textit{Total loss.}
The final objective function is the weighted sum of these components:
\begin{equation}
\mathcal{L}_{\text{total}} = \lambda_p \mathcal{L}_P + \lambda_c \mathcal{L}_{conf} + \lambda_i \mathcal{L}_I,
\end{equation}
where $\lambda_p, \lambda_c, \lambda_i$ are balancing hyperparameters.

\subsection{Training strategy.}
Our EasyPolar pipeline is implemented using PyTorch and trained on an NVIDIA RTX4090 GPU. We train the two-stage pipeline for 2000 epochs in total with an initial learning rate of $1 \times 10^{-4}$, which is decayed by a factor of 0.5 every 100 epochs. The Adam optimizer~\citep{kingma2014adam} is used for optimization with $\beta_1=0.9$ and $\beta_2=0.999$. The training set consists of 200 synthetic scenes. During training, we randomly sample $256 \times 256$ patches from full-resolution images and apply random horizontal flipping for data augmentation. Regarding the hyperparameters in the loss function, we empirically set the balancing weights to $\lambda_p = 1.0$, $\lambda_c = 0.5$, and $\lambda_i = 1.0$. The gradient loss weight is set to $\lambda_g = 0.1$. For the focus weights dealing with imbalance, we set $\alpha = 5.0$ for DoP and $\beta = 2.0$ for confidence supervision.

\section{Experiments}

\begin{table*}[t]
    \centering
    \caption{Quantitative comparisons on synthetic data and real data. The comparisons involve EasyPolar, hardware-based method (DoFP + PIDSR~\citep{zhou2025pidsr}) and generative baseline (Polaranything~\citep{zhang2025polaranything})}
    \label{tab:SyntheticData}
    \resizebox{1.\textwidth}{!}{
    \begin{tabular}{llcccccc}
        \toprule
         & \multirow{2}{*}{Method} & $I_0$ & $I_{45}$ & $S_1$ & $S_2$ & $\theta$ & $\rho$ \\ && PSNR$\uparrow$/SSIM$\uparrow$ & PSNR$\uparrow$/SSIM$\uparrow$ & PSNR$\uparrow$/SSIM$\uparrow$ & PSNR$\uparrow$/SSIM$\uparrow$ & MAE$\downarrow$ & PSNR$\uparrow$/SSIM$\uparrow$ \\
         \midrule
         \multirow{3}{*}{Syn Data} & DoFP camera + PIDSR~\citep{zhou2025pidsr} & 41.39 / 0.9788 & 41.00 / 0.9787 & 50.84 / 0.9953 & 52.20 / 0.9963 & 5.97 & 32.80 / 0.9328 \\
         & Polaranything~\citep{zhang2025polaranything} & 27.86 / 0.8023 & 28.63 / 0.8263 & 27.75 / 0.6580 & 28.83 / 0.6935 & 45.32 & 17.82 / 0.1707 \\
        & Ours & \textbf{41.42} / \textbf{0.9965} & \textbf{42.15} / \textbf{0.9972} & \textbf{51.02} / \textbf{0.9955} & \textbf{53.16} / \textbf{0.9965} & \textbf{3.28} & \textbf{33.15} / \textbf{0.9483} \\
        \midrule
        \multirow{3}{*}{Real Data} & DoFP camera + PIDSR~\citep{zhou2025pidsr} & 35.22 / 0.9148 & 35.04 / 0.9103 & 40.30 / 0.9687 & 44.42 / 0.9779 & 11.97 & 26.27 / 0.7403 \\
         & Polaranything~\citep{zhang2025polaranything} & 34.61 / 0.9034 & 36.07 / 0.9171 & 34.89 / 0.8807 & 35.88 / 0.8910 & 44.76 & 19.71 / 0.2314 \\
        & Ours & \textbf{40.85} / \textbf{0.9711} & \textbf{44.12} / \textbf{0.9901} & \textbf{40.52} / \textbf{0.9691} & \textbf{44.51} / \textbf{0.9802} & \textbf{10.04} & \textbf{26.54} / \textbf{0.7455} \\
        \bottomrule
        \end{tabular}
    }
    \vspace{-3mm}
\end{table*}

\begin{figure*}[t]
\centering
    \includegraphics[width=1.0\linewidth]{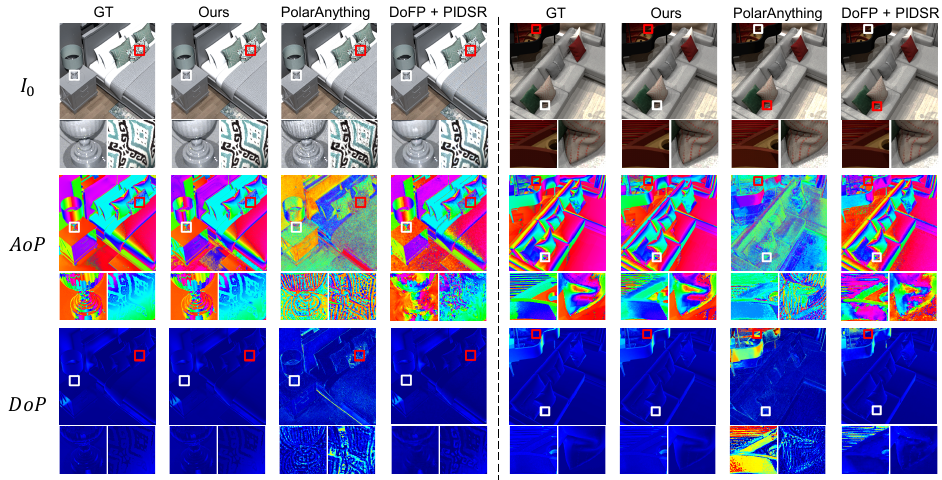}
    \caption{Qualitative comparison on synthetic data between EasyPolar, PolarAnything~\citep{zhang2025polaranything}, and the DoFP camera results processed by a state-of-the-art polarization demosaicing and SR method (PIDSR~\citep{zhou2025pidsr}).}
    \label{fig:qualitative_syn}
    % \vspace{-2mm}
\end{figure*}
\subsection{Experimental Setup}\label{sec:setup}
\noindent\textbf{Baselines.} To comprehensively assess performance, we compare EasyPolar against two representative baselines in both synthetic and real-world experiments:
\textbf{Hardware-based Baseline (DoFP+PIDSR):} This baseline represents the standard snapshot sensor workflow. For synthetic data, we simulate the imaging process by mosaicking the ground-truth polarization channels ($0^\circ, 45^\circ, 90^\circ, 135^\circ$) to generate CPFA raw images. For real-world data, we utilize raw captures from a commercial DoFP camera. In both cases, the raw polarized images are processed by PIDSR~\citep{zhou2025pidsr}, a state-of-the-art joint demosaicing and super-resolution algorithm, to recover the full-resolution polarization parameters.
\textbf{Generative Baseline (PolarAnything):} We evaluate PolarAnything~\citep{zhang2025polaranything}, a recent diffusion-based estimator. We feed it the same RGB inputs used in our method to assess its zero-shot performance.

\noindent\textbf{Metrics.} We employ standard quantitative metrics to evaluate the reconstruction quality across different polarization representations. Specifically, we calculate the Peak Signal-to-Noise Ratio (PSNR) and Structural Similarity (SSIM) for the polarized intensities ($I_0, I_{45}$), Stokes parameters ($S_1, S_2$), and DoP. For AoP, we report the Mean Angular Error (MAE) to measure the angular deviation from the ground truth.

\subsection{Evaluation on Synthetic Data}\label{sec:synthetic_exp}
\noindent\textbf{Dataset generation.} We use the Mitsuba renderer\footnote{\url{https://www.mitsuba-renderer.org/}} to generate a synthetic multi-view polarization dataset tailored to our experimental setup. To ensure diverse scene content, we use the HSSD dataset\footnote{\url{https://3dlg-hcvc.github.io/hssd/}} and select a subset of scenes. For each scene, we render images from different regions and vary camera parameters, including focal length and baseline, to improve generalization. In total, we generate 630 images, with 600 used for training and the remaining 30 for testing. This synthetic dataset provides a diverse and controlled environment for evaluating our method.

\noindent\textbf{Results analysis.} Quantitative comparisons are reported in \tref{tab:SyntheticData}, where our method significantly outperforms the two compared baselines across all metrics. Visual comparisons are further presented in \fref{fig:qualitative_syn}. As observed, the demosaicing-based approach suffers from aliasing artifacts and color crosstalk, particularly in high-frequency texture regions due to the inherent loss of spatial resolution. In contrast, EasyPolar effectively leverages the stereo context to recover sharp and accurate polarization details. Consequently, our method exhibits significantly reduced noise levels and superior reconstruction fidelity compared to the baselines.

\begin{figure*}[t]
\centering
    \includegraphics[width=1.0\linewidth]{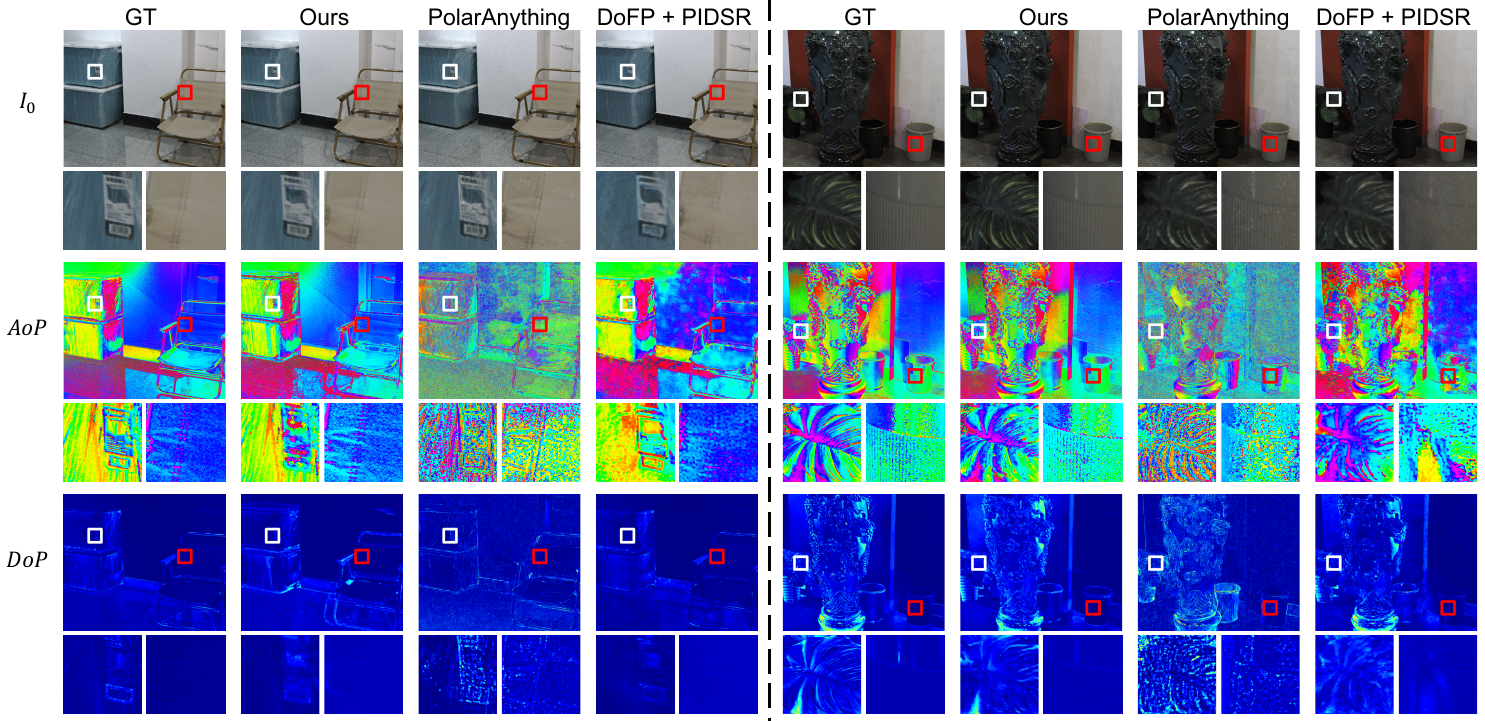}
    \caption{Qualitative comparison on real-world data between EasyPolar, PolarAnything~\citep{zhang2025polaranything}, and the DoFP camera results processed by a state-of-the-art polarization demosaicing and SR method (PIDSR~\citep{zhou2025pidsr}).}
    \label{fig:qualitative_real}
    % \vspace{-2mm}
\end{figure*}

\begin{table*}[t]
    \centering
    \caption{Quantitative evaluation results of ablation study.}
    \label{tab:ablation}
    \resizebox{1.\textwidth}{!}{
    \begin{tabular}{lcccccc}
        \toprule
        \multirow{2}{*}{Method} 
        & $I_0$ 
        & $I_{45}$ 
        & $S_1$ 
        & $S_2$ 
        & $\theta$ 
        & $\rho$ \\
        & PSNR$\uparrow$/SSIM$\uparrow$ 
        & PSNR$\uparrow$/SSIM$\uparrow$ 
        & PSNR$\uparrow$/SSIM$\uparrow$ 
        & PSNR$\uparrow$/SSIM$\uparrow$ 
        & MAE$\downarrow$ 
        & PSNR$\uparrow$/SSIM$\uparrow$ \\
        \midrule

        (1) intensity regression
        & 24.97 / 0.9250
        & 26.57 / 0.9415
        & 24.96 / 0.8989
        & 26.57 / 0.8943
        & 29.09
        & 12.22 / 0.2217 \\

        (2) w/o AoP trigonometric encoding
        & 40.05 / 0.9658
        & 43.14 / 0.9824
        & 39.84 / 0.9651
        & 43.25 / 0.9782
        & 11.24
        & 26.01 / 0.7402 \\

        (3) w/o geometry encoder
        & 40.26 / 0.9684
        & 43.57 / 0.9851
        & 40.08 / 0.9682
        & 43.81 / 0.9795
        & 10.87
        & 25.85 / 0.7357 \\

        (4) w/o confidence gated
        & 39.80 / 0.9648
        & 43.06 / 0.9887
        & 39.62 / 0.9665
        & 43.48 / 0.9794
        & 12.10
        & 25.43 / 0.7294 \\

        (5) w/ $L_1$ loss
        & 40.10 / 0.9693
        & 43.43 / 0.9878
        & 39.75 / 0.9672
        & 43.62 / 0.9813
        & 10.62
        & 25.69 / 0.7307 \\

        \textbf{Full Model (Ours)}
        & \textbf{40.85} / \textbf{0.9711} 
        & \textbf{44.12} / \textbf{0.9901} 
        & \textbf{40.52} / \textbf{0.9691} 
        & \textbf{44.51} / \textbf{0.9802} 
        & \textbf{10.04} 
        & \textbf{26.54} / \textbf{0.7455} \\
        \bottomrule
        \end{tabular}
    }
\end{table*}

\begin{figure*}[t]
\centering
    \includegraphics[width=1.0\linewidth]{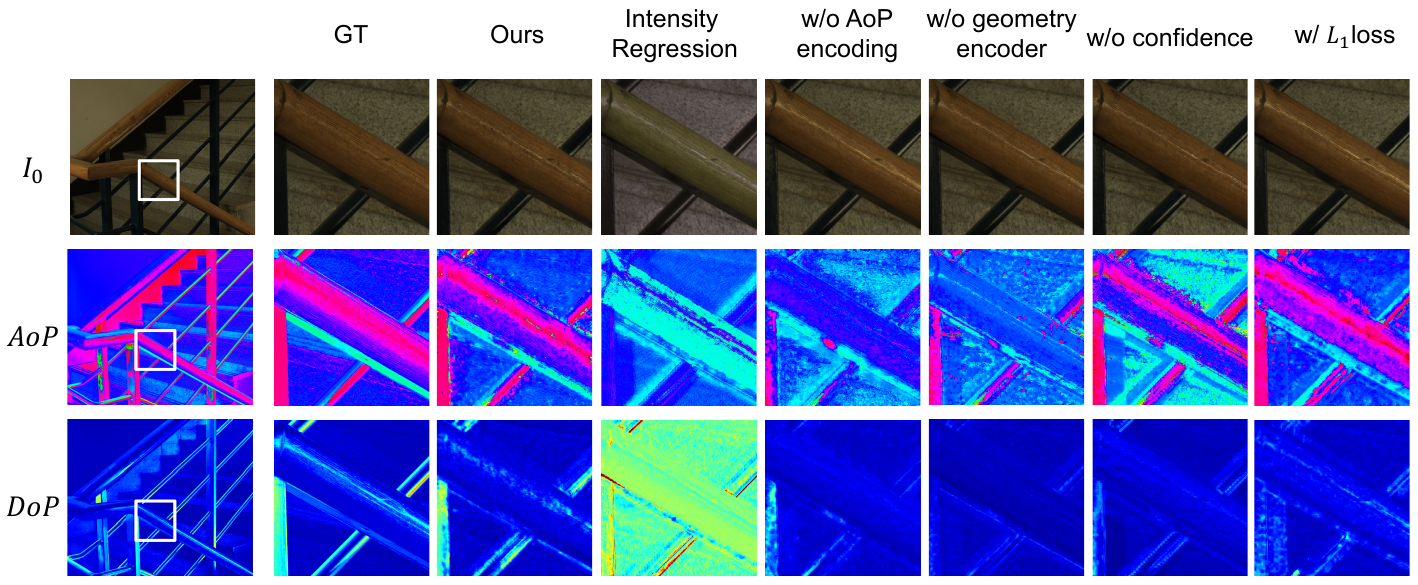}
    \caption{Qualitative results of the ablation study. The results demonstrate that our EasyPolar yields the most accurate polarization details and geometric structures compared to the ablated versions.}
    \label{fig:ablation}
\end{figure*}

\subsection{Evaluation on Real-world Data}\label{sec:real_exp}
\noindent\textbf{Hardware prototype and acquisition.} To validate the proposed sensing principle in the wild, we constructed a synchronized imaging system consisting of three high-resolution RGB cameras mounted with a narrow horizontal baseline (\fref{fig:setup}). The left and right cameras are equipped with fixed linear polarizers at $0^\circ$ and $45^\circ$, respectively, providing the core angular constraints required for single-shot polarization recovery. The middle camera adopts a dual-mode configuration to bridge the gap between training and deployment. During data collection, a motorized rotatable polarizer is temporarily attached to the middle camera to capture a sequence of images at $0^\circ, 45^\circ, 90^\circ$, and $135^\circ$. This time-multiplexed acquisition provides high-fidelity ground-truth polarization parameters for training supervision. During inference, the motorized polarizer is removed, allowing the middle camera to serve as a high-SNR unpolarized reference ($I_{un}^M$) in a single exposure. This hardware--software co-design ensures that the system remains physically sufficient for resolving Stokes vectors while maximizing light throughput via the unpolarized central view.

\noindent\textbf{System calibration and validation.} Since polarimetric calculations rely on precise intensity differences, global radiometric variations among cameras can induce significant errors in the reconstructed polarization states. To mitigate these hardware-induced discrepancies, we perform radiometric calibration prior to data collection. We use a standard X-Rite ColorChecker under uniform illumination to align the white balance and exposure levels of the side cameras to the central reference view. Assuming that the ColorChecker patches are Lambertian and unpolarized, we compute a global color correction matrix (CCM) and an intensity scaling factor for each camera. Applying these corrections ensures that intensity measurements across views are photometrically consistent, thereby isolating physical variations caused solely by polarization and geometry. Furthermore, the narrow-baseline configuration reduces occlusion and parallax, facilitating accurate cross-view correspondence for the alignment module.

\begin{figure*}[t]
\centering
    \includegraphics[width=1.0\linewidth]{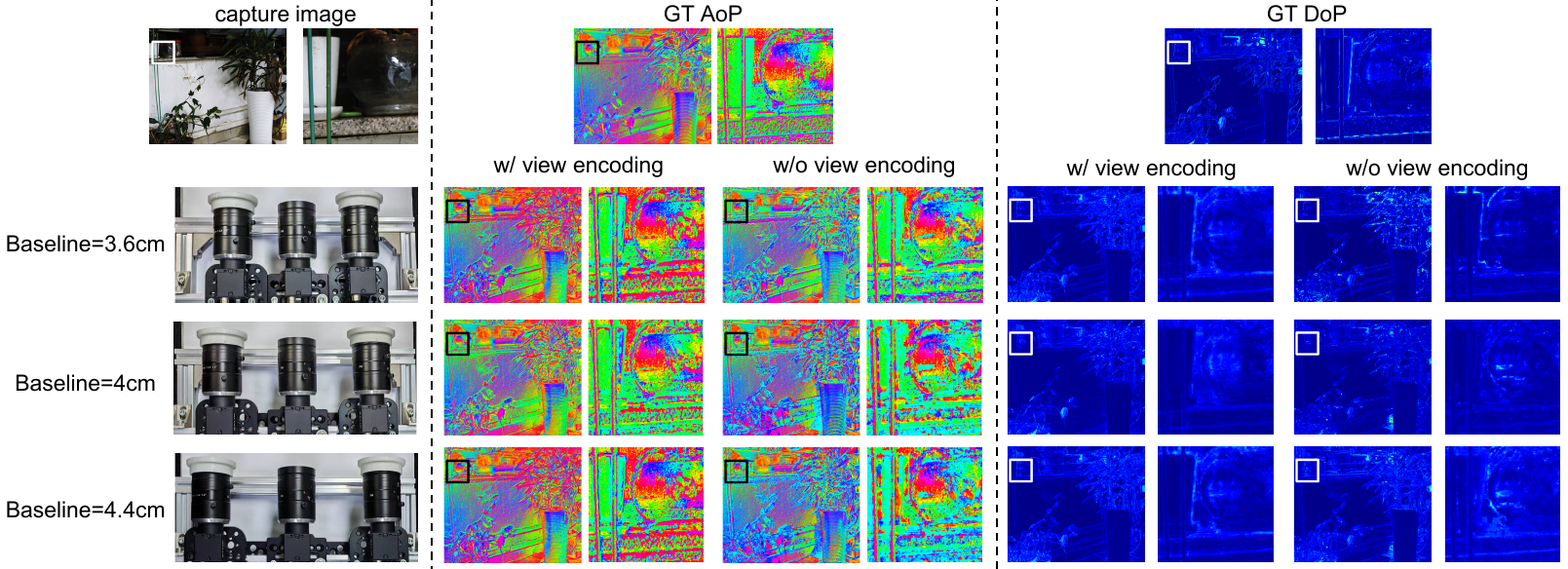}
    \caption{Robustness to camera baseline variations. We evaluate our method under three different baseline distances. With the proposed view encoding, our full framework achieves accurate AoP and DoP estimates with minimal variance across baselines, whereas the version without view encoding fails to handle larger disparities, leading to noticeable artifacts.}
    \label{fig:baseline_ablation}
    % \vspace{-2mm}
\end{figure*}

\begin{figure*}[t]
\centering
    \includegraphics[width=1.0\linewidth]{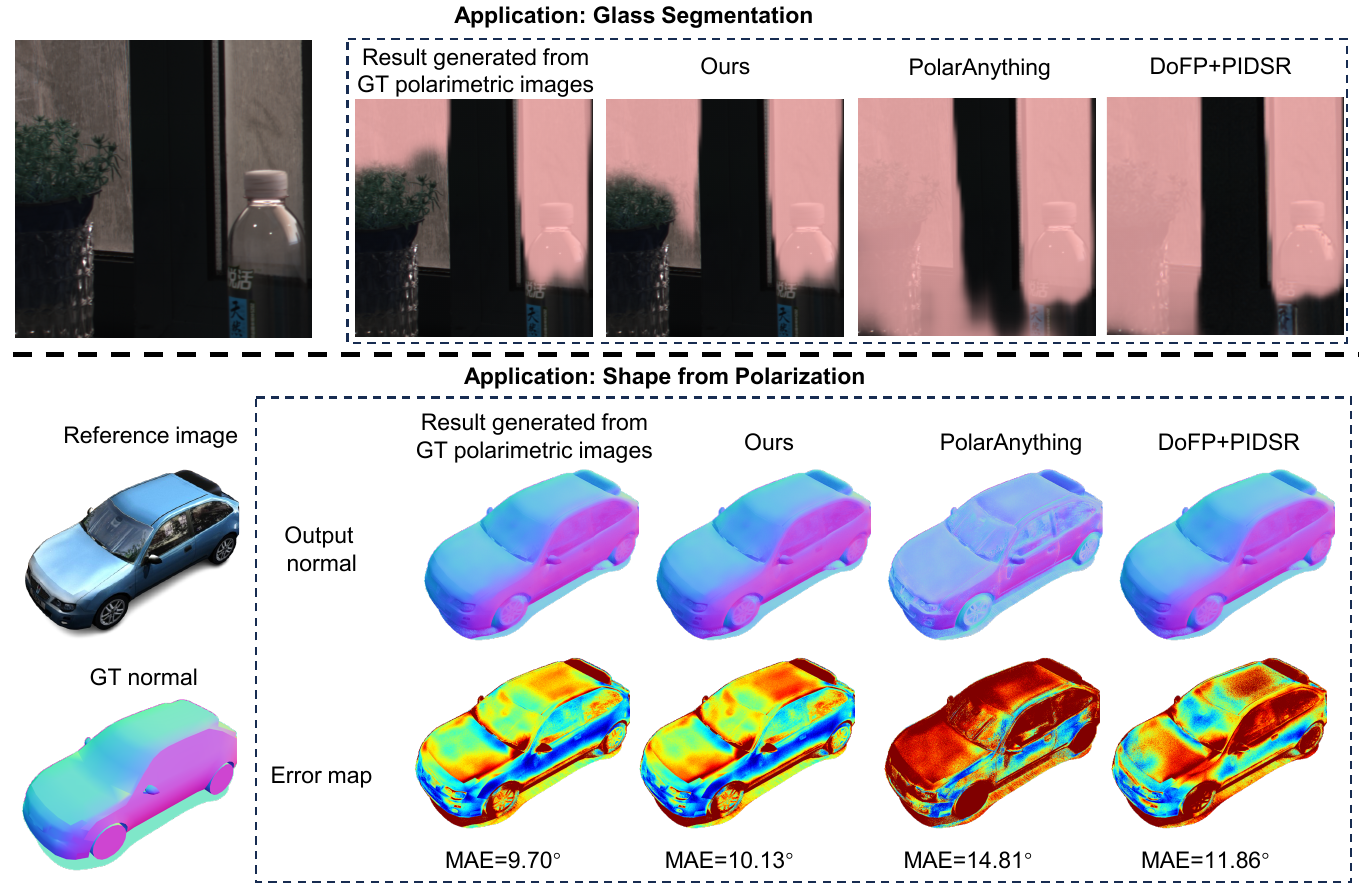}
    \caption{Qualitative comparison on polarization-enhanced downstream tasks. (\textbf{Top}) Glass segmentation: Using our estimated polarization images, a standard glass segmentation method produces cleaner masks that align more closely with the reference (derived from GT polarimetric images) compared to other baselines. (\textbf{Bottom}) Shape from polarization: The mean angular error (MAE) of surface normals, estimated using the method of~\cite{lyu2024sfpuel}, is annotated below each result. The MAE of EasyPolar is the most consistent with that of the GT polarization, demonstrating our method's superior fidelity in preserving physical polarization cues.}
    \label{fig:applications}
    % \vspace{-2mm}
\end{figure*}

\noindent\textbf{Qualitative evaluation.} We evaluate the proposed method in real scenes. \Fref{fig:qualitative_real} presents the visual comparison against the baselines. For the hardware baseline, despite the use of specialized sensors, the results suffer from inherent spatial resolution. Even when enhanced by PIDSR, the reconstructed AoP and DoP maps exhibit noticeable noise and blurring artifacts. Regarding the generative baseline, PolarAnything fails to produce physically plausible estimations for complex scene geometries. This failure stems from a significant domain gap, as it is primarily trained on object-centric data with simple backgrounds. In contrast, EasyPolar demonstrates robust generalization. By effectively fusing stereo intensity cues with physical constraints, our method yields significantly cleaner edges and superior noise suppression, accurately recovering polarization states for both foreground objects and intricate backgrounds.

\subsection{Ablation Study}

To validate the effectiveness of the proposed components in EasyPolarNet, we conduct comprehensive ablation experiments on the synthetic test set. Quantitative results are summarized in \tref{tab:ablation}, and visual comparisons are provided in \fref{fig:ablation}. We analyze the impact of key design choices, including the regression target space, the AoP encoding strategy, and the contribution of specific functional modules, as follows:

\noindent\textbf{(1) Direct polarization prediction.} We compare our direct prediction scheme against indirectly learning polarization via intensity regression. As shown in \tref{tab:ablation}, the indirect approach leads to catastrophic failure, yielding extremely high error rates. Qualitatively, as observed in \fref{fig:ablation}, the results are completely erroneous and structurally distorted. This confirms that calculating polarization from regression-based intensities is highly unstable due to sensitive differential ratios, whereas direct prediction is essential for stable convergence.

\noindent\textbf{(2) Trigonometric AoP encoding.} Scalar regression suffers from ``wrap-around'' discontinuities at the periodic $[0, \pi)$ boundary. Without our continuous $(\sin 2\theta, \cos 2\theta)$ encoding, the model exhibits severe artifacts and large angular errors, particularly in regions near the phase wrap. Our encoding eliminates these discontinuities by aligning the numerical loss with the geometric distance.

\noindent\textbf{(3) Geometry encoding.} Following Fresnel equations, explicit surface normal inputs act as critical physical priors. Removing this module causes the network to fail in decoupling geometric versus material variations. Consequently, the reconstruction quality degrades significantly, showing regional errors on curved surfaces where Fresnel effects are dominant.

\noindent\textbf{(4) Confidence gating.} We evaluate the confidence mechanism by removing its estimation branch and gating module. Without a localized reliability metric, the network fails to suppress erroneous signals from occluded or misaligned regions. This leads to increased noise and degraded boundary sharpness, proving that confidence-based control is vital for robust cross-view polarimetric fusion.

\noindent\textbf{(5) Loss function design.} We replace our tailored objective with a standard $L_1$ loss to validate its effectiveness. Due to the inherent sparsity of polarization signals, the uniform $L_1$ loss tends to cause underestimation. Our design, incorporating high-polarization reweighting and gradient constraints, is essential to counteract data imbalance and recover fine details.

\subsection{Generalization across Stereo Baselines}
To evaluate the capability of our view encoding module in handling geometric variations, we conduct experiments using three distinct stereo baselines: 3.6~cm, 4.0~cm, and 4.4~cm. We compare the full EasyPolar network against an ablated version without the Plücker view encoding, as shown in \fref{fig:baseline_ablation}.

The results demonstrate that, by incorporating view encoding, our method reconstructs consistent and accurate polarimetric information while maintaining sharp structures across all baseline configurations. Conversely, removing this module leads to significant misalignments and a loss of high-frequency details as the baseline increases. This ablation validates that explicit view encoding is critical for stabilizing multi-view fusion, allowing the system to adapt flexibly to different hardware baselines.

\section{Applications}
To demonstrate that EasyPolar serves as a robust foundation for downstream polarization-based vision applications, we evaluate it on two distinct tasks: Glass Segmentation and Shape from Polarization (SfP). For both applications, we compare three processing pipelines: (1)``EasyPolar$\rightarrow$Task'' (Ours) , (2) ``PolarAnything$\rightarrow$Task'' (Generative baseline), and (3) ``DoFP+PIDSR$\rightarrow$Task'' (Hardware baseline).

\noindent\textbf{Application I: Glass segmentation.} We first use glass segmentation~\citep{mei2022glass} on real-world data to test the physical consistency of the reconstructed features. Glass regions typically exhibit strong polarization signatures that are difficult to detect from intensity alone. We compare the segmentation masks generated from the polarization maps of the three pipelines against a segmentation result derived directly from ground-truth polarized images. As shown in the top row of \fref{fig:applications}, while the hardware and generative baselines exhibit erroneous segmentation results in certain regions, our result is significantly more accurate. It aligns most closely with the reference, effectively avoiding the misclassification observed in competing methods and demonstrating superior reliability.

\noindent\textbf{Application II: Shape from polarization.} Next, we validate the geometric accuracy using Shape from Polarization on the synthetic dataset, where ground-truth geometry is available. We adopt the method of~\cite{lyu2024sfpuel} to recover surface normals from the estimated polarization parameters. Visual comparisons are shown in the bottom row of \fref{fig:applications}. As indicated by the annotated metrics, our result achieves a lower Mean Angular Error (MAE) compared to both the hardware and generative baselines, demonstrating its superiority in providing high-fidelity physical cues for 3D reconstruction.

\section{Conclusion}
In this paper, we presented EasyPolar, a complete polarimetric imaging framework comprising a custom-built multi-camera prototype and a tailored reconstruction algorithm. This hybrid design effectively circumvents the resolution loss of conventional DoFP sensors while maintaining the single-shot efficiency lacking in DoT systems. By synergizing explicit geometric priors with confidence-aware physical guidance, our method enables robust fusion of multi-view inputs. Extensive experiments on both synthetic and real-world datasets demonstrate that EasyPolar significantly outperforms state-of-the-art hardware-based and learning-based baselines. Furthermore, validation on downstream polarimetric vision tasks confirms the physical fidelity of our results, highlighting the potential of our system as a reliable acquisition solution for high-precision physics-based vision applications.

\noindent\textbf{Limitations.}
While EasyPolar delivers high-quality results, its performance inherently depends on the accuracy of cross-view geometric correspondence. Despite our confidence-guided fusion, obtaining precise disparity remains a common challenge, particularly in complex scenarios. Moreover, current stereo matching algorithms are primarily optimized for homogeneous systems; extending EasyPolar to heterogeneous devices (e.g., modern smartphones with wide-angle and telephoto lenses) highlights the lack of robust heterogeneous disparity estimation methods, which we identify as a promising direction for future work.

% --- Declarations Section for IJCV / Springer ---
\section*{Statements and Declarations}

%% 1. 资助信息 (Funding)
\paragraph{Funding}
This work was supported by Beijing Municipal Science \& Technology Program No. Z251100007125021, Hebei Natural Science Foundation Project No. 242Q0101Z, Beijing-Tianjin-Hebei Basic Research Funding Program No. F2024502017, National Natural Science Foundation of China (Grant No. 62472044, U24B20155, U23B2052).

%% 2. 利益冲突声明 (Competing Interests)
\paragraph{Competing Interests}
The authors have no relevant financial or non-financial interests to disclose.

%% 3. 数据可用性声明 (Data Availability)
\paragraph{Data Availability}
The source code and datasets used in this study will be made publicly available upon formal acceptance of the manuscript.

\bibliographystyle{spbasic}
\bibliography{egbib}

\end{document}